\documentclass[11pt]{article}

\usepackage[final]{acl}

\usepackage{times}
\usepackage{latexsym}
\usepackage[T1]{fontenc}
\usepackage[utf8]{inputenc}
\usepackage{microtype}
\usepackage{inconsolata}
\usepackage{graphicx}
\usepackage{booktabs}      % \toprule \midrule \bottomrule
\usepackage{amsmath}
\usepackage{multirow}
\usepackage{makecell}
\usepackage{xcolor}
\usepackage{float}   % [H] for the in-section agreement/flip figures
\usepackage{tikz}
\usetikzlibrary{shapes,arrows.meta,positioning,fit,backgrounds,calc}
\usepackage[skins,breakable]{tcolorbox}

\definecolor{clrPrompt}{HTML}{B0CA97} % green  — Prompt Engineering (baseline)
\definecolor{clrSFT}{HTML}{E8A087}    % salmon — SFT (generative)
\definecolor{clrCls}{HTML}{88C4C8}    % teal   — ClsHead (discriminative)
\definecolor{clrEns}{HTML}{D1BC8A}    % gold   — ensemble / majority vote
\definecolor{clrPCS}{HTML}{B7A0C9}    % purple — PCS scope (positive-class)

\newcommand{\teamname}{N\"urnberg NLP}

\newcommand{\TODO}[1]{\textcolor{red}{\textbf{[TODO:~#1]}}}


\newcommand{\fone}{\mathit{F1}}   % F1 with full-size 1, matching \fcv/\ftest
\newcommand{\fcv}{\mathit{F1}_{\mathrm{cv}}}
\newcommand{\fcvtop}{\mathit{F1}_{\mathrm{cv}}^{\mathrm{top3}}}   % per-branch top-3-fold mean (selection signal)
\newcommand{\fcvens}{\mathit{F1}_{\mathrm{cv}}^{\mathrm{ens}}}   % out-of-fold ensemble majority macro-F1 (internal system score)
\newcommand{\ftest}{\mathit{F1}_{\mathrm{test}}}

\newcommand{\rankoverall}{first}

\newcommand{\fcta}{89.56}
\newcommand{\fdbo}{71.63}
\newcommand{\fvio}{54.84}
\newcommand{\fdef}{83.02}

\newcommand{\nmaster}{23{,}829}
\newcommand{\nsynth}{2{,}634}

\title{\teamname{} @ GermEval Shared Task 2026: Harmful Content Detection in German Social Media through Error-Independent LLM Voters}

\author{Philipp Steigerwald \quad Eric Rudolph \quad Jens Albrecht \\
  Technische Hochschule N\"urnberg Georg Simon Ohm \\
  \texttt{\{philipp.steigerwald,\,eric.rudolph,\,jens.albrecht\}@th-nuernberg.de}}

\begin{document}
\maketitle

% ===========================================================================
% Abstract — jargon-free, most abstract level, ~12 lines.
% ===========================================================================
\begin{abstract}
% Opener reframed (Jens review, item 1): societal harm first, the task that
% addresses it second, the technical challenge (imbalance) third.
Harmful content in German social media does real-world damage, from calls to action to criminal defamation.
The GermEval~2026 shared task scores its detection in four subtasks.
The technical challenge is a severe class imbalance.
The harmful classes are rare and share surface language with the dominant majority class, yet under macro-$\fone$ they decide the score.
The decisive lever is then not a stronger single model but error independence.
This insight becomes a per-subtask nine-voter ensemble spanning three orthogonal
axes: LLM, training method and class scope.
Selected mainly on internal cross-validation, the system reaches macro-$\fone$
of \fcta{} (C2A), \fdbo{} (DBO), \fvio{} (VIO) and \fdef{} (DEF) on the hidden
test set, placing \rankoverall{} on all four subtasks.
\end{abstract}

% ===========================================================================
\section{Introduction}
\label{sec:intro}
Harmful content in social media causes real-world damage, from a call to action (C2A) through attacks on the democratic basic order (DBO) and violence-related content such as glorification (VIO) to defamation (DEF) crossing into a criminal offence \citep{zufall2019legal}.
The GermEval~2026 shared task \citep{felser2026germeval} scores these as four subtasks by macro-$\fone$, on anonymised German tweets from a right-wing extremist network.

% Problem motivation + research question (Jens review, item 5): the rare
% classes are hard to tell apart, an example grounds it, the question follows.
Each subtask carries a dominant class of $87$--$97\%$ (Table~\ref{tab:data}), yet the metric weights every class equally, so the rare harmful classes decide the score.
What makes them hard is that they read alike.
On DBO, sharp criticism of the government borders illegal agitation against the state.
The violence classes of VIO differ in pragmatic function rather than wording.
The research question is thus how to separate rare classes that share their surface.

In this shared task's first edition \citep{felser2025germeval}, the best systems leaned on ensembling \citep{koepcke2025nymera} and minority-class augmentation \citep{papadatos2025synthetic}, two of the particularly successful approaches.
The system presented here builds on both --- voting and augmenting --- and rests on voter error independence.
The hypothesis, revisited in the analysis, is that voters of different LLMs and training methods rarely fail on the same tweet, so a plain majority stays right where single voters go wrong \citep{dietterich2000ensemble}.

% ---------------------------------------------------------------------------
% CONSTRAINT: Figure 1 (hero) MUST sit at the TOP of the RIGHT column (col 2)
% on page 1. Declared HERE (after the intro paragraphs have filled column 1)
% with [t] so LaTeX forces it to the next column top = col 2 top. This is the
% ONLY forced float in the paper; every other float uses [h].
% Do not move it earlier, or [t!] lands at the top of column 1 (top-left).
% ---------------------------------------------------------------------------
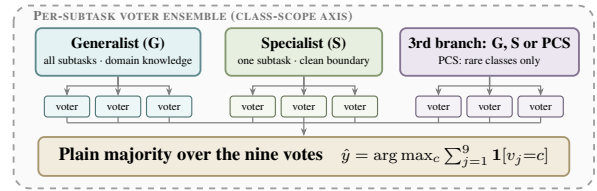
\begin{figure}[t!]
\centering
\resizebox{\columnwidth}{!}{%
\begin{tikzpicture}[
    % --- Hero figure, compact 3-row layout (height matters on page 1): -------
    % scope boxes -> nine voter chips -> gold majority bar. Deliberately
    % text-light: which model/folds fill a branch is prose (§4), not figure
    % text. Cool pastels for the branches (teal G, green S, purple 3rd), gold
    % for the ensemble bar (clrEns = the palette's ensemble colour).
    sysbox/.style={rectangle, rounded corners=3pt, line width=0.9pt,
        minimum width=2.7cm, minimum height=0.82cm, align=center,
        inner ysep=2pt, inner xsep=4pt, font=\fontsize{7.5}{9}\selectfont\bfseries},
    % voterchip = one vote: three per branch (its top-3 folds), tinted in the
    % branch colour so the nine voters visibly group into the three branches.
    voterchip/.style={rectangle, rounded corners=1.5pt, line width=0.55pt,
        minimum width=0.78cm, minimum height=0.36cm, align=center,
        inner sep=1pt, font=\fontsize{5.4}{6.4}\selectfont},
    major/.style={rectangle, rounded corners=3pt, line width=0.9pt,
        align=center, font=\footnotesize\bfseries, inner ysep=4pt, inner xsep=10pt},
    edge/.style={-{Stealth[length=4pt]}, line width=0.6pt, black!50},
    bus/.style={line width=0.6pt, black!50},
]
\begin{scope}[on background layer]
  \draw[draw=black!38, dashed, rounded corners=8pt, line width=0.9pt,
        fill=black!2] (-5.0, 1.30) rectangle (5.0, -1.85);
\end{scope}
% --- container label, vertically centred between the dashed top edge (1.30)
% --- and the top edge of the scope boxes (0.91) -> mid 1.105.
\node[font=\fontsize{6.5}{8}\selectfont\bfseries, text=black!55,
      anchor=west] at (-4.8, 1.105)
      {\textsc{Per-subtask voter ensemble (class-scope axis)}};
% --- three scope branches. The third slot is filled per subtask: a second G
% --- or S on the binary subtasks, the PCS where rare classes exist (DBO/VIO);
% --- only the DBO winner keeps the PCS. Kept generic here on purpose.
\node[sysbox, draw=clrCls!70!black, fill=clrCls!22] (b1) at (-3.2, 0.50)
    {Generalist (G)\\[-1.5pt]{\fontsize{5.6}{7}\selectfont\mdseries all subtasks · domain knowledge}};
\node[sysbox, draw=clrPrompt!75!black, fill=clrPrompt!30] (b2) at (0, 0.50)
    {Specialist (S)\\[-1.5pt]{\fontsize{5.6}{7}\selectfont\mdseries one subtask · clean boundary}};
\node[sysbox, draw=clrPCS!72!black, fill=clrPCS!26] (b3) at (3.2, 0.50)
    {3rd branch: G, S or PCS\\[-1.5pt]{\fontsize{5.6}{7}\selectfont\mdseries PCS: rare classes only}};
% --- nine visible voters: each branch fans out into its three voter chips
% --- (tinted in the branch colour), so 3 branches x 3 voters is explicit ----
\foreach \b/\xc/\clr in {b1/-3.2/clrCls, b2/0/clrPrompt, b3/3.2/clrPCS}{
  \foreach \dx in {-0.88, 0, 0.88}{
    \node[voterchip, draw=\clr!60!black, fill=\clr!14] at ({\xc+\dx}, -0.44) {voter};
    % orthogonal wiring: down from the box, horizontal, down into the chip
    \draw[edge] (\b.south) |- ({\xc+\dx}, -0.085) -- ({\xc+\dx}, -0.26);
    \draw[bus]  ({\xc+\dx}, -0.62) -- ({\xc+\dx}, -0.72);
  }
}
% --- collector bus -> plain majority over the nine votes --------------------
\draw[bus] (-4.08, -0.72) -- (4.08, -0.72);
\node[major, draw=clrEns!80!black, fill=clrEns!28] (ens) at (0, -1.30)
    {Plain majority over the nine votes\quad{\fontsize{8}{9}\selectfont\mdseries
     $\hat{y}=\arg\max_{c}\sum_{j=1}^{9}\mathbf{1}[v_j{=}c]$}};
\draw[edge] (0, -0.72) -- (ens.north);
\end{tikzpicture}%
}
\caption{Per-subtask ensemble: three scope branches, three voters each, plain
majority decides.}
\label{fig:arch}
\end{figure}
% Camera-ready (Review 1, R1-2): the two closing intro paragraphs compressed —
% voter details and the chosen combinations live in §4 — and "branch" is
% glossed here (Reviews 1+2: the term must precede Figure 1 and §4).
% Project rule kept: SFT spelled out, generative/discriminative explained at
% first mention; model names stay out of the intro.
Five LLMs enter the pool, four between 7 and 14 billion parameters and one at 128 billion.
An optimised prompt \citep{yang2023opro} sets the untrained baseline, while the voters are fine-tuned under 4-bit QLoRA \citep{dettmers2023qlora} in two modes, generative supervised fine-tuning (SFT), which emits the label as text, and a discriminative classification head (ClsHead) reading it from the hidden state.
Every voter trains on one of three class scopes, a generalist (G) over all subtasks, a specialist (S) over its own subtask and, for DBO and VIO only, a positive-class specialist (PCS) over their positive classes alone.
Guided by five-fold cross-validation, the three best model--method--scope combinations supply three voters each --- a branch --- so nine voters per subtask decide every tweet by plain majority (Figure~\ref{fig:arch}).
Such voting sharpens the blur between semantically close classes \citep{steigerwald2026psydefdetect} and the system ranks \rankoverall{} on all four subtasks, contributing the voter architecture and the augmentation data.

% Camera-ready (Review 1, R1-1): structure overview closes the introduction —
% deliberate exception to the no-section-refs-in-the-intro rule.
Section~\ref{sec:related} reviews related work, Section~\ref{sec:data} describes the data and Section~\ref{sec:system} the system.
Section~\ref{sec:results} reports the hidden-test results, Section~\ref{sec:analysis} analyses the winning ensembles and Section~\ref{sec:conclusion} concludes.

% ===========================================================================
\section{Related Work}
\label{sec:related}
Related work falls into three strands: the German task family, the ensemble and fine-tuning techniques and the prior voter ensemble it extends.

% 2.1 Camera-ready (Review 1, R1-3): widened beyond the GermEval family
% (GermEval 2018, HASOC, survey); koepcke/papadatos moved to 2.2 (Review 2,
% R2-2), where they fit thematically. The corpus itself is described in §3.
\subsection{Harmful Content in German Tweets}
Beyond this task family, harmful-content detection has a broad history, from the GermEval~2018 shared task on offensive language \citep{wiegand2018germeval} to the multilingual HASOC track \citep{mandl2019hasoc}, surveyed by \citet{fortuna2018survey}.
The subtasks here extend a line of German resources, from the DeTox corpus \citep{demus2022detox} to a human-annotated dataset of criminally relevant hate speech \citep{kums2025botox}.
DEF follows the legal scheme of \citet{zufall2019legal}.
In the prior edition \citep{felser2025germeval}, a (Q)LoRA sweep over German and multilingual models took C2A with ModernGBERT-1B and VIO with the German-native LL\"aMmlein-7B \citep{wunderle2025superglber}.
Other systems fine-tuned encoders under heavy imbalance \citep{bullard2025moderngbert} or engineered classical features \citep{schwichtenberg2025features}, while prompting alone fell short of the fine-tuned top systems \citep{ludwig2025prompting}.

% 2.2 Camera-ready (Reviews 1+2, R1-3/R2-2): ensembles for abusive-language
% detection in general, alternative imbalance treatments (SMOTE, loss-side,
% survey) and the resource angle; koepcke/papadatos now cited HERE.
\subsection{Ensembles and Fine-Tuning under Imbalance}
The ensemble rests on error-independence theory \citep{dietterich2000ensemble}, under which diverse voters that make uncorrelated mistakes are corrected by majority voting.
Ensembles have a long record in abusive-language detection, from deep-learning ensembles for hate speech \citep{zimmerman2018hate} to bagged BERT models for aggression identification \citep{risch2020bagging}.
In the prior edition, the DBO winner soft-voted three BERT variants \citep{koepcke2025nymera}.
Such encoder ensembles cost far less compute than LLM voters.
Class imbalance is met on the data side or the loss side \citep{henning2023survey}.
Data-side treatments range from classical minority oversampling \citep{chawla2002smote} to LLM-generated synthetic minority tweets, which \citet{papadatos2025synthetic} validated on this task family.
This system uses both sides, synthetic minority tweets and a focal loss \citep{lin2017focal} that shifts the loss towards the hard minority examples.
Training uses 4-bit QLoRA \citep{dettmers2023qlora}.
The prompting baseline follows Optimization by PROmpting (OPRO) \citep{yang2023opro}, where an LLM iteratively proposes prompts scored against a development split.

% 2.3 Own prior work: the multi-axis voter ensemble and its CV5 discipline; PCS is new.
% CV-folds-as-ensemble + held-out estimate goes back to Krogh & Vedelsby (1995).
\subsection{Multi-Axis Voter Ensembles}
The voter pool design follows a prior multi-axis voter ensemble \citep{steigerwald2026psydefdetect}, which paired a generalist over the full label set with specialists over a reduced one and drew its voters from class granularity, training method and LLM.
Its cross-validation folds doubled as voter pool and performance estimate \citep{krogh1995ensembles}, a discipline the selection here inherits.
The positive-class specialist is the new step, trained only on non-majority examples.

% ===========================================================================
\section{Data}
\label{sec:data}
% Subtask definitions follow the organiser class definitions verbatim (they are
% the SFT prompts, Appendix E); DEF legal scheme = zufall2019legal (cited in RW).
% Overview-paper citation added for camera-ready (Review 1, R1-5; also cited
% at the task's first mention in the introduction, user pick B).
The shared task \citep{felser2026germeval} provides anonymised German tweets from a right-wing extremist network, labelled for four independent subtasks.
% Acronyms spelled out at the subtask definitions (Jens review, item 6).
\textbf{C2A} (call to action) asks whether a tweet calls its readers to a specific action, criminal or not --- the early signal of mobilisation.
\textbf{DBO} (democratic basic order) grades a tweet's stance towards the free democratic basic order in four classes. Its difficulty is the boundary between legitimate \textit{criticism} of the government and illegal \textit{agitation} or \textit{subversive} calls to overthrow it.
% Paper prose spells "propensity" correctly (user decision 2026-07-04, no
% footnote); the RELEASED label string and therefore all submissions, prompts
% (Appendix E) and code keep the organisers' spelling "prospensity".
\textbf{VIO} (violence) separates how violence enters a tweet in six classes, from the author's own readiness (\textit{propensity}) through \textit{call2violence}, \textit{support} and \textit{glorification} to \textit{other}.
\textbf{DEF} (defamation) asks whether a tweet meets the criminal-law standard of defamation (\S\S185--187 StGB), not whether it merely feels offensive.
% The majority class described for DBO and VIO (Review 2, R2-3/R2-4; user
% pick B: one shared sentence instead of two half-sentences).
On DBO and VIO the label \textit{nothing} marks the majority class, tweets that show none of these phenomena.
% Jens item 7, reworked (Philipp, 2026-07-10): this sentence no longer mentions
% augmentation at all — its first §3 appearance is the motivated "To counter
% this imbalance" sentence below; the intro already announced the approach
% ("builds on both", minority-class augmentation as proven prior recipe).
The training release holds \nmaster{} tweets and Table~\ref{tab:data} lists the per-label counts.
% Overlap measured on data/clean/master.jsonl (2026-07-04): 75.3% of the
% 23,829 tweets carry labels for >=2 of the four official subtasks
% (8,602 for two, 8,745 for three, 595 for all four).
Three quarters of these tweets are labelled for two or more subtasks, so the per-subtask counts in Table~\ref{tab:data} add up to well beyond \nmaster{}.
DEF is an order of magnitude smaller than the other three subtasks and on DBO and VIO the decisive positive classes fall to a fraction of a percent.

% Per-subtask labels, real training counts and realised minority-class
% augmentation. Caption <= 2 lines; the imbalance is visible in the counts
% (e.g. VIO nothing 15,901 vs glorification 27). aug numbers from master.jsonl +
% all_train_aug source=augmented; budget eq:aug (realised falls slightly short).
% Last column = class share AFTER augmentation: (n_c + aug_c) / (N + aug_total)
% with N/aug_total = 15839/0 (C2A), 15853/544 (DBO), 16431/1535 (VIO), 3263/555 (DEF).
\begin{table}[h]
  \centering
  \small
  \setlength{\tabcolsep}{3pt}
  \begin{tabular*}{\columnwidth}{@{\extracolsep{\fill}}l l r r r r@{}}
    \toprule
    \textbf{Subtask} & \textbf{Label} & \textbf{Train} & \textbf{Prior} & \textbf{+Aug} & \textbf{Prior$_{\text{+aug}}$} \\
    \midrule
    \multirow{2}{*}{C2A}
      & false & 14{,}405 & 90.9\% & ---    & 90.9\% \\
      & true  & 1{,}434  & 9.1\%  & ---    & 9.1\%  \\
    \midrule
    \multirow{4}{*}{DBO}
      & nothing    & 13{,}818 & 87.2\% & ---    & 84.3\% \\
      & criticism  & 1{,}839  & 11.6\% & ---    & 11.2\% \\
      & agitation  & 143      & 0.9\%  & +400 & 3.3\%  \\
      & subversive & 53       & 0.3\%  & +144 & 1.2\%  \\
    \midrule
    \multirow{6}{*}{VIO}
      & nothing       & 15{,}901 & 96.8\% & ---    & 88.5\% \\
      & propensity    & 74       & 0.5\%  & +222 & 1.6\%  \\
      & call2violence & 197      & 1.2\%  & +571 & 4.3\%  \\
      & support       & 102      & 0.6\%  & +306 & 2.3\%  \\
      & glorification & 27       & 0.2\%  & +81  & 0.6\%  \\
      & other         & 130      & 0.8\%  & +355 & 2.7\%  \\
    \midrule
    \multirow{2}{*}{DEF}
      & false & 2{,}848 & 87.3\% & ---    & 74.6\% \\
      & true  & 415     & 12.7\% & +555 & 25.4\% \\
    \bottomrule
  \end{tabular*}
  \caption{Per-subtask label counts. Prior is the label's share before,
  Prior$_{\text{+aug}}$ after augmentation.}
  \label{tab:data}
\end{table}

\label{sec:aug}
% "we generated" (Jens review, item 8): makes clear the synthetic tweets are
% ours, added on top of the official release.
To counter this imbalance, we generated \nsynth{} synthetic tweets with GPT-5.4 for the minority classes only (Table~\ref{tab:data}, +Aug), prompting it with the full class taxonomy and five real examples per class (Appendix~\ref{sec:app-augprompt}).
% Budget formula inline (Philipp, 2026-07-10): as a display equation it
% dangled at the column bottom; it is never referenced, so it flows in-text.
The per-class budget is $\mathrm{aug}_c = \min\!\left(\text{target}-n_c,\; 3\,n_c\right)$ with $\text{target}=1000$, a round figure well above the rare-class counts yet far below the majority.
A rare class is thus lifted toward a thousand examples unless the cap at three times its real count $n_c$ binds first, which keeps synthetic tweets at most $75\%$ of a class.
% Why C2A gets no synthetic tweets (Review 1, R1-4): its positive class already
% exceeds the target, so the budget formula yields zero.
C2A receives no synthetic tweets, as its positive class already holds $1{,}434$ examples and sits above the target.
A few generations fail a sanity check for malformed output or duplicate text and are dropped, so the realised counts in Table~\ref{tab:data} fall a little short of this budget and are not topped up by regeneration.
Synthetic data enters the training folds only, validation and test stay human-annotated.

% ===========================================================================
\section{System}
\label{sec:system}
The system is a voter ensemble in which nine voters per subtask decide every tweet's label by a plain majority vote (Figure~\ref{fig:arch}).
% Branch defined at first use (Reviews 1+2, R1-6): the term appears in Figure 1
% and §4.1/§4.4 before the formal §4.5 walk-through, so it is introduced here;
% §4.5 now only back-references it (user pick B, no duplication).
The nine voters come as three \emph{branches}: a branch is three voters that share one LLM, training method and class scope and differ only in their training folds.
The idea behind it is that different models make different errors, so the joint vote is meant to sharpen the semantically fuzzy boundaries between the classes.
The construction of these nine voters rests on a small set of concepts.
The following subsections introduce the components in ascending order --- the LLMs, the adaptation methods, the class scopes and the cross-validation protocol they train under --- then the architecture, voter selection and voting rule that assemble them into the ensemble.
The natural starting point is the choice of LLM.

\subsection{Models}
\label{sec:models}
The model axis spans five LLMs of different families, sizes and training regimes, so it injects genuinely different inductive biases into the pool.
\textbf{min} (Ministral-8B-Instruct)~\citep{mistral2024ministral} and \textbf{phi4} (Phi-4-14B)~\citep{abdin2024phi4} are general-purpose instruction-tuned models of 8B and 14B parameters, chosen as strong all-rounders that led our prior multi-axis voter ensemble on a comparable German classification task \citep{steigerwald2026psydefdetect}.
\textbf{LLäMmlein-7B}~\citep{pfister2025llammlein} is German-native, so its pretraining sits closest to the tweets.
It enters in two variants, a base model (\textbf{LLaM-B}) and a chat-tuned model (\textbf{LLaM-I}).
% minR (Ministral-3-8B-Reasoning) was dropped from the paper (2026-06-29): it never
% entered a deployed ensemble and is in none of the six submissions. Its heatmap CV5
% was inflated (150-sample RSFT-eval estimates), its ClsHead head degenerated to a
% single class, and its SFT specialist was only a weak, partially val-banked
% diversifier, so the CV5-based selection never picked it. Removing it keeps the
% model roster to the ones that actually carry a voter.
A single much larger model is added to see how far raw parameter count carries and how much it adds over the compact models: \textbf{Mis128B} (Mistral-Medium-3.5, 128B)~\citep{mistral2025medium}, which also serves as the optimiser for prompt search.
All models run 4-bit quantised and the full tweet always enters without truncation.
% GPU-memory figures (Jens item 9; ALL probe-verified 2026-07-10, jobs
% 181602-04/181688-89/181605/181692: bank load path, 4-bit NF4). Measured p99
% peaks: LLaM-B 5.0 / min 6.1 / phi4 9.4 GiB; phi4 4096-token overflow 10.3 GiB
% on one H200 (explains the ml0 11-GB OOM); mis128 on H200 (16x768/4x4096):
% load 65.9, p99 73.7, overflow 77.0 GiB -> the 70-85 GB claim holds. Parallel
% sums (GiB): VIO ~52, DEF ~71, C2A ~261, DBO ~480 -> 50-75 and 240-530 hold.
Quantised this way, a compact voter runs inference in 5--10\,GB of GPU memory and the 128B model in 70--85\,GB.
The voters are independent, so an ensemble answers either sequentially, where its largest voter sets the peak, or fully in parallel at the sum of its nine voters --- 50--75\,GB for nine compact voters and 240--530\,GB for the compositions that carry 128B branches.
How such an LLM turns into a classifier is the next choice, the adaptation method.

\subsection{Adaptation Methods}
\label{sec:methods}
% Opener restructured (Jens review, item 10): first name the three methods,
% the shared training details moved below the three paragraphs.
Three adaptation methods are studied, in increasing order of supervision.
Prompt engineering is the baseline, while the two trained modes, generative supervised fine-tuning (SFT) and a discriminative classification head (ClsHead), make up the ensemble.

\paragraph{Prompt engineering (OPRO).}
Every model gets a system prompt optimised by OPRO \citep{yang2023opro}.
In this loop the Mis128B model writes candidate prompts and the target model tries each one, classifying a class-balanced sample of 10--20 training tweets per class.
Prompts that score a higher macro-$\fone$ seed the next round and the best one becomes the model's baseline prompt.
Prompting needs no training and sets the bar the trained methods have to clear.

\paragraph{SFT (generative).}
The model is fine-tuned to emit the class label as text from a German task prompt with the full category definitions (Appendix~\ref{sec:app-prompts}).

\paragraph{ClsHead (discriminative).}
A two-layer classification head on the last-token hidden state of the QLoRA-adapted model is trained with focal loss \citep{lin2017focal} and inverse-frequency class weights, which fight the imbalance directly.

% Cross-method sentences pulled out of the ClsHead paragraph (pass-1 re-check
% fix 7): they describe both modes, not the head.
Both trained modes use 4-bit NF4 QLoRA \citep{dettmers2023qlora} on every linear projection and their training hyperparameters are listed in Appendix~\ref{sec:app-hparams}.
The two modes make different errors on different tweets and the vote exploits these complementary error profiles.
The remaining choice of which training data a voter sees is its class scope.

\subsection{Class Scope: Generalist, Specialist and Positive-Class Specialist}
\label{sec:pcs}
The scope axis sets which classes a voter trains on and carries the main idea of the system, with each scope meant to contribute a different strength.

\paragraph{Generalist (G).}
A generalist trains jointly on all subtasks in a multi-task setup.
The shared signal is meant to build broad domain knowledge --- patterns common to the subtasks that no single one provides alone.
% "training modes (SFT and ClsHead)" re-established here (Jens review, item 11).
% The switch is a GENERALIST property (specialists train on one subtask and
% need none) — the generalist is the subject so this cannot be misread as an
% architecture-wide feature (Philipp, 2026-07-10).
Evaluation still runs one subtask at a time, so the generalist carries a subtask switch in both training modes (SFT and ClsHead).

% --- Generalist mechanism (v5, Philipp 2026-07-10): BOTH pipelines carry the
% --- per-subtask prompt (verified in train_clshead.py build_input_text);
% --- dashed = per-subtask selector. ClsHead-G additionally routes through one
% --- head per subtask. Shared five-slot grid so the rows mirror.
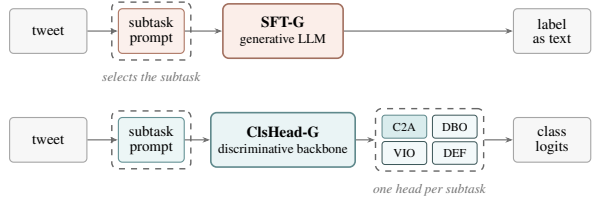
\begin{figure}[h]
\centering
\resizebox{\columnwidth}{!}{%
\begin{tikzpicture}[
    io/.style={rectangle, rounded corners=2pt, line width=0.55pt, align=center,
        minimum width=1.35cm, minimum height=0.78cm, fill=black!3, draw=black!35,
        font=\fontsize{7.4}{8.8}\selectfont},
    bb/.style={rectangle, rounded corners=3pt, line width=0.8pt, align=center,
        minimum width=2.1cm, minimum height=0.95cm,
        font=\fontsize{7.6}{9}\selectfont},
    swin/.style={rectangle, rounded corners=2pt, line width=0.55pt, align=center,
        minimum width=1.15cm, minimum height=0.78cm,
        font=\fontsize{7.4}{8.8}\selectfont},
    hd/.style={rectangle, rounded corners=1.5pt, line width=0.5pt, align=center,
        minimum width=0.78cm, minimum height=0.34cm,
        font=\fontsize{6}{7.2}\selectfont},
    sw/.style={draw=black!60, dashed, rounded corners=3pt, line width=0.7pt,
        inner sep=3pt},
    ar/.style={-{Stealth[length=3.8pt]}, line width=0.6pt, black!55},
    lbl/.style={font=\fontsize{6.6}{8}\selectfont\itshape, text=black!55,
        align=center},
]
% ---- row 1: SFT-G -- the prompt selects, the label comes back as text ------
\node[io] (t1) at (-5.1, 0.95) {tweet};
\node[swin, draw=clrSFT!75!black, fill=clrSFT!16] (p1) at (-3.3, 0.95)
    {subtask\\[-1.5pt]prompt};
\node[bb, draw=clrSFT!70!black, fill=clrSFT!18] (bb1) at (-1.0, 0.95)
    {\textbf{SFT-G}\\[-1pt]{\fontsize{6.6}{7.8}\selectfont generative LLM}};
\node[io] (o1) at (3.7, 0.95) {label\\[-1.5pt]as text};
\draw[ar] (t1) -- (p1);
\draw[ar] (p1) -- (bb1);
\draw[ar] (bb1) -- (o1);
\node[sw, fit=(p1)] (sw1) {};
\node[lbl, anchor=north] at ([yshift=-2pt]sw1.south) {selects the subtask};

% ---- row 2: ClsHead-G -- same prompt selection plus one head per subtask ---
\node[io] (t2) at (-5.1, -0.95) {tweet};
\node[swin, draw=clrCls!75!black, fill=clrCls!16] (p2) at (-3.3, -0.95)
    {subtask\\[-1.5pt]prompt};
\node[bb, draw=clrCls!70!black, fill=clrCls!18] (bb2) at (-1.0, -0.95)
    {\textbf{ClsHead-G}\\[-1pt]{\fontsize{6.6}{7.8}\selectfont discriminative backbone}};
% 2x2 grid of per-subtask heads, target head highlighted
\node[hd, draw=clrCls!70!black, fill=clrCls!32] (hC) at (1.12, -0.72) {C2A};
\node[hd, draw=clrCls!45!black, fill=clrCls!10]  (hD) at (2.00, -0.72) {DBO};
\node[hd, draw=clrCls!45!black, fill=clrCls!10]  (hV) at (1.12, -1.18) {VIO};
\node[hd, draw=clrCls!45!black, fill=clrCls!10]  (hF) at (2.00, -1.18) {DEF};
\node[sw, fit=(hC)(hD)(hV)(hF)] (sw2) {};
\node[io] (o2) at (3.7, -0.95) {class\\[-1.5pt]logits};
\draw[ar] (t2) -- (p2);
\draw[ar] (p2) -- (bb2);
\draw[ar] (bb2) -- (sw2);
\draw[ar] (sw2) -- (o2);
\node[sw, fit=(p2)] (swp2) {};
\node[lbl, anchor=north] at ([yshift=-2pt]sw2.south) {one head per subtask};
\end{tikzpicture}%
}
\caption{Both modes select the subtask by its prompt (dashed); \mbox{ClsHead-G}
adds one head per subtask.}
\label{fig:generalist}
\end{figure}

% Generalist mechanism, final (Philipp, 2026-07-10): ONE model in both modes,
% BOTH trained with one prompt per subtask (verified in train_clshead.py
% build_input_text — task name + categories are prepended per sample).
% "backbone" is reserved for ClsHead (backbone + 4 heads); the SFT generalist
% is just the fine-tuned LLM itself.
In both modes one model serves all subtasks, trained with one prompt per subtask (Appendix~\ref{sec:app-prompts}) and the prepended prompt selects the subtask at inference (Figure~\ref{fig:generalist}).
The SFT generalist is the plain fine-tuned LLM and emits the selected subtask's label as text.
The ClsHead generalist splits into a shared backbone (the QLoRA-adapted LLM) and four classification heads, one per subtask, reading only the selected subtask's head.

\paragraph{Specialist (S).}
A specialist trains only on the data of a single subtask.
The narrowed training signal is meant to fit that subtask's label set more tightly, though in practice the specialist performs on a par with the generalist.
What it mainly contributes is a different error profile the vote can use.

\paragraph{Positive-class specialist (PCS).}
A positive-class specialist is a specialist trained without the majority class.
It applies only to DBO and VIO, whose several positive classes must be told apart, not to the binary C2A and DEF.
Removing the dominant class is meant to shift the model's errors elsewhere and free its capacity for the fine distinctions between the positive classes.
% Reading verified against fig_embeddings.pdf (mis128-base-fulltrain, job
% 179424): the positive classes sit intermixed in the grey majority cloud.
Even the untrained Mis128B, the strongest LLM in the pool, places these rare classes close together in its raw representation, % Appendix prefix added (Jens review, item 13) so readers do not hunt for
% the figure in the main body.
with no clean boundary off the shelf (Figure~\ref{fig:emb}, Appendix~\ref{sec:app-emb}).
The positive-class specialist is trained to separate exactly these.
How these voters are trained and scored is the task of the cross-validation protocol.

\subsection{Cross-Validation and Evaluation}
\label{sec:cv}
% "class-balanced" spelled out as stratified (Review 2, R2-5); implementation
% is a stratified 5-fold split over the combined labels of all subtasks.
The master training set is split into five folds for five-fold cross-validation (CV5).
The split is stratified over the labels of all subtasks, so each fold mirrors the overall label distribution and no tweet appears in more than one fold.
A branch is trained five times, each run on four folds and scored on the held-out fifth, giving five per-fold macro-$\fone$ scores $\fcv$.
The mean of the three best, $\fcvtop$, is its selection signal and names the three voters it deploys.
A PCS branch is scored only on its positive classes, so its $\fcvtop$ sits on a higher scale and is not directly comparable to a full-label branch's.

% Paragraph reordered (Jens review, item 14): mechanics first, the
% "doubles as" conclusion follows from them.
An ensemble's internal estimate stands in for the withheld $\ftest$ yet comes from the training data its voters trained on, so each tweet is scored out-of-fold, only by a voter that held out its fold and never saw it.
Keeping all five fold models per branch, not only the deployed three, gives every tweet one such leakage-free vote per branch.
A tweet $i$ with text $x_i$ comes from fold $k(i)$, so in each branch $b$ it is predicted by the fold model $m_b^{k(i)}$ that held that fold out.
The out-of-fold label $\hat{y}_i$ is the majority vote (mode) over the three branches,
\begin{equation}
  \hat{y}_i=\operatorname{mode}\bigl\{\,m_b^{k(i)}(x_i)\,\bigr\}_{b=1}^{3},
  \label{eq:oof}
\end{equation}
whose macro-$\fone$ against the gold labels over the whole training set is $\fcvens$.
Cross-validation thus doubles as the pool the voters deploy from and as the ensemble's internal estimate.

That is three votes per tweet, where the deployed top-three folds instead cast nine on the hidden test.
$\fcvens$ is only an estimate, a workaround for the withheld test labels, that guided the build alongside $\fcvtop$.
$\ftest$ is the final result, listed against $\fcvens$ per submission in Table~\ref{tab:allsubs}.
With voters, branches and their scores in place, the parts assemble into the architecture.

\subsection{Architecture}
\label{sec:architecture}
Three terms build the system from a single vote up.
A \textbf{voter} is one model, fine-tuned in one of the two modes on one class scope and trained on four of the five folds, then evaluated on the held-out fifth that names it.
A \textbf{branch}, introduced above, deploys the three highest-scoring of its five fold voters, so it casts three votes.
% Grid corrected (camera-ready): three scopes exist only on DBO/VIO, the
% binary subtasks have no PCS.
Five models and two modes give $5\times2$ branches per scope, over three scopes on DBO and VIO and two on the binary subtasks, which have no positive-class specialist.
An \textbf{ensemble} is three branches, nine voters in all, that decide every tweet by plain majority (Figure~\ref{fig:arch}).
% Why exactly three branches (Review 2, R2-6; user decision: spell the balance
% out, odd count as a side effect, axis/fold coverage as capability --- the VIO
% winner is single-mode and skips fold 3, so no "always" claim).
An ensemble never repeats a branch, and in every deployed composition the branches differ in at least the LLM or the training method, so the intended diversity sits between branches while the three folds within a branch add redundancy against single-model noise.
Three branches give a composition room to span two LLMs, both training methods and two class scopes at once, and their nine voters can cover all five folds.
More branches would pull weaker candidates from the pool and shrink each voter's share of the vote, a fourth branch cutting it from a ninth to a twelfth.
Nine votes are also an odd count, so the binary subtasks cannot tie, a side effect rather than a design goal.
Six- and twelve-voter variants were also submitted and the nine-voter blueprint won every subtask.
How an ensemble's three branches are chosen is the selection problem addressed next.

\subsection{Voter Selection}
\label{sec:selection}
Choosing an ensemble means choosing its three branches per subtask, and cross-validation decides it.
% The class-balance check is the one non-CV signal (pass-1 re-check fix 2):
% named here so the abstract's "mainly" holds and Limitations is not the
% reader's first encounter with it.
One check runs outside it: a candidate whose predicted share of the majority class on the test tweets falls far below the training prior is discarded before submission.
A first stage ranks the candidate branches by $\fcvtop$ and favours those whose per-fold scores correlate weakly, since branches that fail on different folds make more independent errors.
A second stage scores whole compositions by $\fcvens$ (Eq.~\ref{eq:oof}) and keeps the best.
% The earlier claim that the deployed folds always cover all five was false
% for the VIO winner (folds 0,1,2,4; verified in analysis/voter_agreement.json).
% Top-3 by score is the rule, full coverage a preference (pass-1 re-check fix 1).
A branch deploys its three highest-scoring folds, and where compositions score alike the one covering more of the five folds is preferred, which three of the four winners achieve.
The hidden test set plays no part.

The scheme is fixed per subtask family.
For the binary C2A and DEF a composition spans two models, both modes and the two full-label scopes, arranged as 2G\,$\times$\,S or G\,$\times$\,2S.
% Third-branch role on the binary subtasks made explicit (Review 1, R1-7): no
% PCS exists there, so the third branch is a second full-label branch that
% contributes a third error profile.
A positive-class specialist does not exist on a binary subtask, so the third branch is a second generalist or second specialist under a different model or mode, contributing a third error profile rather than a rare-class view.
The fine-grained DBO and VIO add a positive-class specialist as a possible third branch (G\,$\times$\,S\,$\times$\,PCS).
% The 6V/12V sentence moved to §4.5 (R2-6), where the three-branch design is
% motivated — no duplication here.
Every deployed composition is listed in Table~\ref{tab:allsubs} and Tables~\ref{tab:perfold-c2a} to~\ref{tab:perfold-def} list the per-model, per-method ranking signal.
What remains is the rule by which the nine voters decide.

\subsection{Voting Rule}
\label{sec:voting}
The ensemble decision is a plain majority over all nine votes of the three branches, $\hat{y}=\arg\max_{c}\sum_{j}\mathbf{1}[v_j{=}c]$.
Ties are broken towards the lowest class id, the majority class, which on a binary subtask can only happen at an even voter count.
The generalist and specialist branches predict the full label set.
They have seen the majority class far more often than any rare class, so they are expected to be confident on it and to anchor it.
The positive-class specialist never predicts the majority class, yet this stays harmless.
While the full-label voters agree on the majority class it already holds enough votes to win.
The positive-class specialist can only move the decision once they split on a tweet, the uncertain cases where a sharper view of the rare classes helps.
The majority class is thus protected whenever the full-label voters are confident, which is expected to be the common case, while the third voter earns its place on the rare-class calls.
The agreement analysis confirms exactly this mechanic on the test votes.
The scores this system reaches on the hidden test set follow next.

% ===========================================================================
\section{Results}
\label{sec:results}
The metric is per-subtask macro-$\fone$, reported in percent throughout.
The internal estimate $\fcvens$ is the out-of-fold ensemble score (\S\ref{sec:cv}) and $\ftest$ is the official score on the hidden test set (test sizes 2{,}982 C2A, 3{,}165 DBO, 4{,}108 VIO and 577 DEF), whose gold labels are withheld.
Voter codes are derived as model--scope--method, combining the model shorthand (min, phi4, llamB, mis128), the class scope (G, S, PCS) and the method (sft, cls).
For example, llamB-G-cls is the LLaM-B ClsHead generalist.

% The former compact score table (tab:lb) was dropped (Philipp, 2026-07-10):
% every one of its values also sits in the F1test column of tab:allsubs, which
% now carries all per-submission references alone.
The per-subtask ensemble places \rankoverall{} on all four subtasks, with $\ftest$ of \fcta{} (C2A), \fdbo{} (DBO), \fvio{} (VIO) and \fdef{} (DEF) (Table~\ref{tab:allsubs}).
Each winner is a nine-voter plain majority chosen on cross-validation, yet the composition differs by subtask.
The four winning compositions are described per subtask below, followed by the prompting baseline and a summary reading of the submissions.

% Full per-submission, per-subtask internal CV5 (F_cv) vs hidden-test (F_test).
% F_cv/F_test for submissions 1-5 from the post-hoc audit; submission 6 test
% official. VIO is the trap (F_cv rises while F_test falls), DBO PCS over-fires on
% weak models (sub 4) but transfers on the 128B (sub 5), C2A clears 89 only at
% the CV5-max mix (sub 6).
\begin{table*}[t]
  \centering
  \small
  % Long format: one line per (submission, subtask) so every composition sits
  % directly next to its two scores. Compositions verified against
  % score_valbank.py (run1/2/3, reconstructed from the deployed prediction
  % CSVs) and check_submission.py COMPOSITION (subs 4-6, zip-verified).
  % (na) = no-aug. Voter code model-scope-method, defined in the running text.
  \begin{tabular*}{\textwidth}{@{\extracolsep{\fill}}l c l l cc@{}}
    \toprule
    \textbf{Sub.} & \textbf{V} & \textbf{Subtask} & \textbf{Voter composition} & $\fcvens$ & $\ftest$ \\
    \midrule
    \multirow{4}{*}{1} & \multirow{4}{*}{6}
      & C2A & llamB-G-cls + min-G-cls & 84.93 & 89.17 \\
      & & DBO & llamB-G-cls + min-G-cls & 63.56 & 66.89 \\
      & & VIO & llamB-G-cls + min-G-cls & 51.56 & 53.43 \\
      & & DEF & llamB-G-cls + min-G-cls & 79.25 & 78.61 \\
    \midrule
    \multirow{4}{*}{2} & \multirow{4}{*}{9}
      & C2A & llamB-G-cls + min-G-cls + min-S-cls   & 86.00 & 89.08 \\
      & & DBO & llamB-G-cls + min-G-cls + min-S-cls   & 68.41 & 67.15 \\
      & & VIO & llamB-G-cls + min-G-cls + min-S-cls   & 57.21 & \textbf{54.84} \\
      & & DEF & llamB-G-cls + min-G-cls + llamB-S-cls & 81.03 & 80.93 \\
    \midrule
    \multirow{4}{*}{3} & \multirow{4}{*}{12}
      & C2A & llamB-G-cls\,(na) + llamB-G-cls + min-G-cls + min-S-cls\,(na) & 86.08 & 88.78 \\
      & & DBO & llamB-G-cls\,(na) + llamB-G-cls + min-G-cls + min-S-cls\,(na) & 67.23 & 68.67 \\
      & & VIO & llamB-G-cls\,(na) + llamB-G-cls + min-G-cls + min-S-cls\,(na) & 54.58 & 51.64 \\
      & & DEF & llamB-G-cls\,(na) + llamB-G-cls + min-G-cls + min-S-cls\,(na) & 81.25 & 81.61 \\
    \midrule
    \multirow{4}{*}{4} & \multirow{4}{*}{9}
      & C2A & llamB-G-cls + min-S-cls + phi4-S-sft  & 87.01 & 88.39 \\
      & & DBO & min-G-cls + phi4-S-cls + min-PCS-sft  & \textbf{72.42} & 64.37 \\
      & & VIO & min-G-cls + llamB-S-cls + min-PCS-sft & 61.70 & 50.38 \\
      & & DEF & llamB-G-cls + phi4-S-cls + phi4-S-sft & 81.89 & \textbf{83.02} \\
    \midrule
    \multirow{4}{*}{5} & \multirow{4}{*}{9}
      & C2A & phi4-G-cls + mis128-S-sft + phi4-S-sft     & 87.12 & 88.41 \\
      & & DBO & min-G-cls + mis128-S-sft + mis128-PCS-sft  & 71.66 & \textbf{71.63} \\
      & & VIO & min-G-cls + mis128-S-sft + mis128-PCS-sft  & 63.71 & 51.12 \\
      & & DEF & phi4-G-cls + mis128-S-sft + mis128-S-cls   & 81.64 & 80.58 \\
    \midrule
    \multirow{4}{*}{6} & \multirow{4}{*}{9}
      % Sub6/C2A F1cv audited (verify_table2.py): the printed composition
      % yields 87.54, not the earlier 86.92; bold moves here from Sub5.
      & C2A & llamB-G-cls + llamB-S-sft + mis128-S-sft    & \textbf{87.54} & \textbf{89.56} \\
      & & DBO & llamB-G-cls + mis128-S-sft + mis128-PCS-sft & 72.20 & 71.36 \\
      & & VIO & phi4-G-cls + mis128-S-sft + mis128-PCS-sft  & \textbf{66.25} & 51.10 \\
      & & DEF & phi4-G-cls + llamB-S-sft + llamB-S-cls      & \textbf{82.13} & \textbf{83.02} \\
    \bottomrule
  \end{tabular*}
  \caption{Every submission per subtask, internal $\fcvens$ against hidden-test
  $\ftest$ (\%). \textbf{V} = voters per subtask, \textbf{bold} = per-subtask
  best per column, (na) = no augmentation.}
  \label{tab:allsubs}
\end{table*}

\subsection{C2A: Call to Action}
\label{sec:res-c2a}
% C2A F1cv range follows the audited 87.54 (Table 2): 2.19 -> 2.61.
The internal estimate varies by only $2.61$ points across the six submissions and the hidden test set stays flatter still (range $1.17$), not following it, marking a saturated subtask near its ceiling where the composition barely moves the score.
The best reaches \fcta{} with a generalist and two specialists (G\,$\times$\,2S), llamB-G-cls + llamB-S-sft + mis128-S-sft, each branch deployed on its three best cross-validation folds for nine voters in total.

\subsection{DBO: Democratic Basic Order}
\label{sec:res-dbo}
DBO spreads much more widely, across a range of $7.26$ points on the hidden test set and the internal estimate largely carries over, so the composition, not a ceiling, decides the score.
The best reaches \fdbo{} and is the only winner that keeps a positive-class specialist (G\,$\times$\,S\,$\times$\,PCS), min-G-cls + mis128-S-sft + mis128-PCS-sft, each on its three best folds.
The positive-class specialist trains with the majority class removed and DBO is the one subtask where it carries over, the winner moving from $71.66$ internally to $71.63$ on the test.
Two of the three branches ride on the 128B model.
With an 8B specialist the same design over-fired (submission 4, $\ftest{=}64.37$), so the transfer of the positive-class specialist plausibly follows the strength of its LLM.

\subsection{VIO: Violence}
\label{sec:res-vio}
VIO is the hardest subtask and the one where the internal estimate misleads.
It climbs by $14.69$ points across the submissions while the test drifts the other way, staying low within a range of $4.46$.
The best reaches \fvio{} already with the second submission, two generalists and one specialist (2G\,$\times$\,S), all discriminative, llamB-G-cls + min-G-cls + min-S-cls, each on its three best folds.
The diversity here rests on the two LLMs and the scope split rather than the mode.
The variants with a positive-class specialist score higher internally yet do not carry over (Table~\ref{tab:allsubs}), so the winner drops it.

\subsection{DEF: Defamation}
\label{sec:res-def}
Internal estimate and test score rise together across the submissions, so the internal gains carry over most faithfully here.
The composition closes a range of $4.41$ points above the weakest submission and two mirrored compositions reach the best \fdef{}, each a generalist with two specialists (G\,$\times$\,2S) --- llamB-G-cls + phi4-S-cls + phi4-S-sft and phi4-G-cls + llamB-S-sft + llamB-S-cls, the two models swapping roles.
DEF is also the smallest subtask at 577 test tweets.

\subsection{Prompt Engineering versus Fine-Tuning}
\label{sec:ceiling}
Prompt engineering loses to fine-tuning for every model on every subtask (Tables~\ref{tab:perfold-c2a} to~\ref{tab:perfold-def}).
% Deltas recomputed 2026-07-05 from the fresh CV5 pull (Tables 5-8 rows):
% prompt 79.35/35.14/36.02/71.09 vs best fine-tuned 86.78/70.77/59.21/83.79.
% Gains from full-precision values: 7.43/35.63/23.19/12.70 (mean 19.74);
% round to 7/36/23/13, average 20. (The old "8" came from pre-rounded 86.8-79.3.)
Even the 128B, the strongest prompted model, gains $7$ points on C2A, $36$ on DBO, $23$ on VIO and $13$ on DEF in $\fcvtop$ when fine-tuned, an average of $20$.
The gap is largest on the imbalanced multi-class subtasks, where the untrained 128B does not separate the decisive classes in its raw representation (Figure~\ref{fig:emb}, Appendix~\ref{sec:app-emb}) and only fine-tuning reshapes the space enough to tell the rare classes apart.
Prompting is thus the baseline the voter pool clears.

% Method-comparison figure (full width). Caption <= 2 lines.

\subsection{Submission Summary}
\label{sec:summary}
Six submissions were scored and Table~\ref{tab:allsubs} lists each one's composition against its $\fcvens$ and $\ftest$ per subtask.
They form a progression of growing diversity, from a two-generalist cross-model baseline (submission~1) through the second mode and a no-augmentation branch to the positive-class specialist and the 128B model (submissions~5 and~6).
The stronger LLMs carry the hardest subtasks. The 128B specialist lifts DBO to its best $\ftest$ and the cross-model mix tops C2A, while the compact LLaM-B and min models anchor the generalists everywhere.
% "helps consistently" contradicted the all-discriminative VIO winner
% (pass-1 re-check fix 3).
Every winner draws on at least two LLMs and three of the four also mix the two modes, the all-discriminative VIO winner being the exception.

% §5.6 closer sharpened (Philipp, 2026-07-10): per-subtask difficulty named —
% C2A/DEF inherently separable (binary), DBO hard (criticism vs agitation share
% tone and target), VIO hardest (six classes, one violent surface).
Read together, C2A and DEF are inherently the easier subtasks.
Both pose a single binary decision between harmful and harmless content and score accordingly.
DBO is harder.
Its four classes grade the stance towards the democratic order and the decisive boundaries separate legitimate criticism from illegal agitation and subversive calls, three classes that share tone and target.
VIO is the hardest, its five violence classes sharing the violent surface and differing only in pragmatic function (Figure~\ref{fig:emb}, Appendix~\ref{sec:app-emb}).
The next gains therefore lie in understanding these class boundaries rather than in a stronger composition.

% Camera-ready (Review 1, R1-8): runtime and hardware for fine-tuning and
% inference. Training medians from 1,389 train_summary.json (nfs1 + FAU vault),
% pool count 469 fold models / ~4,000 GPUh reproduced against 1,388 summaries;
% latency measured with scripts/analysis/bench_latency.py on the deployment
% code path (analysis/bench_latency.json, mis128 on H200 job 187508).
\subsection{Computational Cost}
\label{sec:cost}
All models train and infer 4-bit quantised on the memory budgets of \S\ref{sec:models}.
% Audited 2026-08-08 (independent agent, train_summary.json + SLURM logs):
% specialist medians 6.2-13.1 h ok; generalist lower bound 14 -> 10 (deployed
% phi4-G-cls 10.2 h); "one H200" was FALSE — 5 of the 9 deployed 128B folds
% trained sharded on TWO H200s (log VRAM fingerprint), 4 on one. Follow-up
% audit of the full job history: all folds started on gpu:1; c2a-f2/f4 hit a
% DETERMINISTIC long-batch CUDA OOM there (fixed seed, f4 twice at step
% 1296/4755; the longest C2A tweet sits in the train data of exactly these
% folds) and retrained on two; dbo-f3/pcs-f1/pcs-f2 had NO OOM and were only
% resubmitted under the newer gpu:2 script after preemption. Batch size is
% identical on 1 and 2 GPUs (bs=4, model-parallel sharding). 4,000 GPUh kept
% as estimate (dedupe wall sum 3,269 h + uncounted multi-GPU/preempted runs).
On one data-centre GPU (NVIDIA A100, L40S or H200) a specialist fold model fine-tunes in a median of $6$--$13$ hours, a multi-task generalist in $10$--$30$ and a positive-class specialist in about $2$.
The 128B specialist takes about $4$ hours per fold under its reduced setup (Appendix~\ref{sec:app-hparams}).
It fits on one H200 at the full batch size, yet a single batch of very long tweets can exceed the card, so part of its folds trained sharded across two H200s.
The full pool behind the selection holds $469$ trained fold models, roughly $4{,}000$ GPU-hours.
Table~\ref{tab:latency} lists the measured inference latency per voter.
A compact ClsHead voter answers on an 11\,GB RTX~2080~Ti in under two seconds per tweet.
SFT voters take longer, as they also generate the label text.
The GPU matters more than the parameter count, the 128B voter answering in $0.9$\,s on one H200.
An ensemble running its nine voters one after another answers a tweet in roughly $6$ to $13$\,s, or at its slowest voter's latency in parallel.

% Median s/tweet from analysis/bench_latency.json (n=64 real test tweets per
% voter, deployment-identical code path). Latency is scope-independent, so the
% rows name model and method only. Caption <= 2 lines.
\begin{table}[h]
  \centering
  \small
  \begin{tabular*}{\columnwidth}{@{\extracolsep{\fill}}llcr@{}}
    \toprule
    \textbf{Model} & \textbf{Method} & \textbf{GPU} & \textbf{s/tweet} \\
    \midrule
    llamB  & ClsHead & RTX 2080 Ti & 0.42 \\
    min    & ClsHead & RTX 2080 Ti & 0.79 \\
    phi4   & ClsHead & RTX 2080 Ti & 1.63 \\
    phi4   & SFT     & RTX 2080 Ti & 2.42 \\
    llamB  & SFT     & RTX 2080 Ti & 2.57 \\
    mis128 & SFT     & H200        & 0.91 \\
    \bottomrule
  \end{tabular*}
  \caption{Measured per-voter inference latency, median seconds per tweet.}
  \label{tab:latency}
\end{table}

% ===========================================================================
\section{Analysis}
\label{sec:analysis}
\label{sec:agreement}
% Numbers from analysis/voter_agreement.json (scripts/analysis/voter_agreement.py).
% Every winner was reproduced EXACTLY (0 diffs) from its nine per-voter test
% predictions via plain majority before any metric below was computed.
% §6 reworked (Jens review, items 18+19): every subsection opens with its
% question and why it matters; the prose keeps only the load-bearing numbers,
% Figures 3/4 and Table 3 carry the rest.
With the submissions scored, the analysis turns to the four winning systems and answers three questions: did the internal estimate predict the test, do the voters disagree as designed and where does the third branch decide?
% DEF ties at 83.02 between two compositions (§5.4); name the one analysed
% (pass-1 re-check fix 6).
Where DEF ties, the earlier of its two compositions is the one analysed.

\subsection{Estimate versus Test}
\label{sec:transfer}
The first question is whether the internal estimate actually predicted the hidden test.
Every ensemble was selected on $\fcvens$, so an estimate that misleads would undermine every choice before it.
The two score columns of Table~\ref{tab:allsubs} give the answer.
% Pearson/gap follow the audited Table-2 cell: 0.9467 -> 0.9468, 3.36 -> 3.34.
Across all 24 rows $\fcvens$ and $\ftest$ correlate at Pearson $r=0.9468$ with a mean absolute gap of $3.34$ points, so the estimate ranked the submissions well before the hidden test was seen.
% Within-subtask r values (0.9021 DEF / 0.2605 DBO / -0.3876 C2A) trimmed from
% the prose (Jens review, item 19); they live in analysis/voter_agreement.json.
Within a single subtask the picture is finer.
DEF is the honest case, estimate and test rising together as the composition strengthens.
DBO tracks as well, apart from one submission whose aggressive positive-class specialist over-fires.
The C2A test scores span only $1.17$ points, so the internal wobble maps to noise.
VIO is the one inversion ($r=-0.61$), the internally strongest submissions gaining rare classes in cross-validation and losing them on the hidden test.

\subsection{Disagreement by Design}
\label{sec:agreement-sub}
\begin{figure}[h]
  \centering
  \includegraphics[width=\columnwidth]{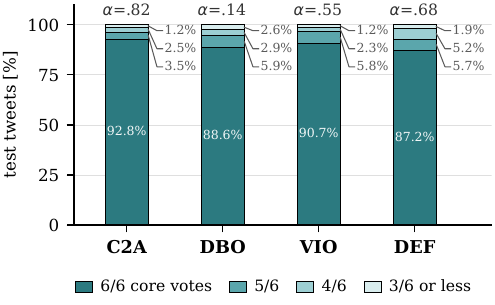}
  % Camera-ready (Review 1, R1-9): sequential teal scale (darker = more
  % agreement), percentages on all segments, small ones via leader lines.
  \caption{Core-voter agreement per winner, darker shades = more agreement,
  nine-voter Krippendorff's $\alpha$ on top.}
  \label{fig:agreement}
\end{figure}

How much the voters actually disagree is the second question.
The system bets on diversity rather than on a stronger single model, so a strong winner's nine voters should not agree too much.
% R2-7 (user pick B): the analysis targets the submitted winners, not copies.
The analysis covers the submitted winners themselves, not retrained copies.
All numbers come from the archived per-voter test predictions behind the scored submissions.
% VIO .551 / DEF .684 trimmed from the prose (item 19); Figure 3 shows all four.
The voters' inter-rater reliability, measured as Krippendorff's $\alpha$ over the nine votes, spans a wide range, from $.825$ on the saturated C2A down to $.142$ on DBO (Figure~\ref{fig:agreement}).
% R2-8 (user pick C): stated as an observation consistent with the data, not
% as a proven finding — the honesty register the reviewer asked to clarify.
The pattern is consistent with disagreement sitting between branches rather than within them, and how wide it opens varies sharply by subtask.

DBO opens it widest.
% Pair-level alphas (.806-.895 within, .025/.022 with PCS, .522 VIO generalists)
% trimmed from the prose (item 19); Figure 3 carries the pairwise detail.
Within each branch the folds agree closely, yet the two branch pairs that involve the positive-class specialist agree barely above zero, which produces the lowest nine-voter agreement of the four winners.
The one winner that keeps a PCS is thus the least agreeing, on the very subtask where the composition moves the score most --- the classical precondition for ensemble gains \citep{dietterich2000ensemble}.
VIO draws its independence from another axis, its lowest branch-pair agreement sitting between the two generalists of different LLMs, so where the training method is shared the model supplies the disagreement.
C2A sits at the other end, its voters agreeing closely throughout, the saturated subtask where the ensemble adds least.
% R2-9 (user pick A): the composition-to-gain pattern named explicitly.
The two ends anchor the pattern: DBO pairs the lowest agreement with the widest composition-driven swing in Table~\ref{tab:allsubs}, C2A the highest agreement with a flat score, and DEF and VIO sit between.
Read from four winners alone this is a marker of a well-composed ensemble, not a proof.

% Core-agreement figure pinned exactly between its two paragraphs ([H]).
% Caption <= 2 lines.

\subsection{Third-Branch Arbitration}
\label{sec:flips-sub}
\begin{figure}[h]
  \centering
  \includegraphics[width=\columnwidth]{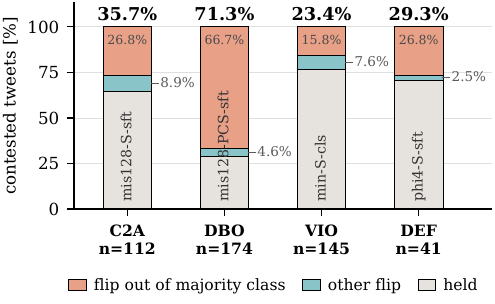}
  % Camera-ready (Review 1, R1-10/R1-11): qualitative colours (the flip
  % categories are unordered) and the caption now says what n counts.
  \caption{Third-branch flips among the $n$ contested tweets per subtask
  (core $\leq$4/6). Black is the total flip rate.}
  \label{fig:flips}
\end{figure}

The third branch decides the close cases the two core branches leave open.
% Unanimity 87-93% -> "nine of ten", contested-zone counts 41-174 trimmed
% (item 19); Figure 4 shows the per-subtask counts.
The six core voters of the first two branches are unanimous on roughly nine of
ten test tweets.
Once at least five of the six agree, the decision is mathematically locked
against the three third-branch votes --- not a single flip occurs there.
The third branch therefore acts only in a small contested zone of $3.5\%$ to
$7.1\%$ of the tweets per subtask, exactly the uncertain cases it was added for
(Figure~\ref{fig:flips}).

% Flip counts (23-36%, 108/124, 30/40, 11/12) trimmed from the prose (item 19,
% Jens' "ab Zeile 544" complaint); Figure 4 splits every flip by direction.
% "a quarter to a third" hid C2A's 36% (pass-1 re-check fix 5).
There it flips between $23\%$ and $36\%$ of the contested decisions on C2A, VIO and
DEF, but $71\%$ on DBO, where nearly all are flips \emph{out of the majority
class} --- the positive-class specialist moves \textit{nothing} to
\textit{criticism}, across the minority boundary that decides macro-$\fone$.
The few \emph{other flips} instead revise a decision the core already placed in
a rare class.
The binary subtasks repeat the pattern, most flips turning \textit{false} into
\textit{true}, so the third branch is the recall lever of the ensemble.
% Third-branch worth quantified (Review 2, R2-10; user pick C, short form).
% Deltas verified against analysis/third_branch_ablation.json
% (scripts/analysis/third_branch_ablation.py): F1cv-ens of each winner with
% and without its third branch; gold labels exist only on cross-validation.
On cross-validation, where gold labels exist, dropping the third branch costs $2.3$ points of $\fcvens$ on C2A, $8.3$ on DBO, $5.7$ on VIO and $3.6$ on DEF.
Part of each figure is the smaller voter count rather than the missing branch's distinct view, so these bound the contribution from above.

% Contested-zone flip figure pinned right after the arbitration paragraph ([H]).
% Caption <= 2 lines.

% ===========================================================================
\section{Conclusion}
\label{sec:conclusion}
% Structure (2026-07-05): three paragraphs --- (1) system + result,
% (2) the three findings, (3) the three open levers.
% "Our system" instead of the team name (Jens review, item 16).
% Research challenge restated up front (Review 2, R2-11, user pick A).
The challenge behind all four subtasks is to separate rare harmful classes that share their surface language with a dominant harmless majority.
Our system answers every tweet of each subtask with a plain majority over nine fine-tuned voters, selected mainly on internal cross-validation across three diversity axes --- LLM, training method and class scope.
It places \rankoverall{} on all four subtasks of the hidden test set, with macro-$\fone$ of \fcta{} (C2A), \fdbo{} (DBO), \fvio{} (VIO) and \fdef{} (DEF).

The decisive lever is voter error independence, not a single stronger model.
Fine-tuning beats every optimised prompt and even the strongest pool model does not separate the rare harmful classes in its raw representation.
% Resource comparison ensemble vs prompted single model (Jens review, item 17):
% compact voters 5-9 GB, prompted 128B 70-85 GB (§4.1) — the all-compact VIO
% winner beats the prompted 128B even on memory; 128B-branch winners do not.
In resources the picture is mixed.
% "less memory" was overstated, the ranges overlap: 50-75 vs 70-85 GB
% (pass-1 re-check fix 4).
Nine compact voters fit in memory comparable to one prompted 128B pass, while 128B branches cost more (\S\ref{sec:models}).
The voters disagree between branches rather than within them.
The third branch earns its votes in the contested zone, most visibly on DBO, where the positive-class specialist pushes undecided tweets into the rare classes.
The internal signal ranked the submissions reliably with one exception.
On VIO the test score does not rise although the internal estimate climbs, so the internal ranking cannot be trusted there.

% Future-work reframe (Philipp, 2026-07-10): two hedged directions — targeted
% disagreement search for a better F1test, and efficiency (smaller models,
% fewer voters per branch) for lower training and inference costs.
Two directions remain for future work.
A targeted search for branches that disagree most while keeping a strong internal signal, guided by the disagreement analysis, could plausibly lift $\ftest$ further.
On the efficiency side, smaller models and perhaps fewer voters per branch could bring the training and inference costs down.

% ===========================================================================
% Local squeeze (Philipp, 2026-07-10): minimal baselinestretch on the back
% matter ONLY (Limitations -> Data Availability) so all content ends on page 8;
% pages 1-7 keep their tuned breaks. Group closes before the references.
{\renewcommand{\baselinestretch}{0.96}\selectfont
\section*{Limitations}
Each submission gives a single hidden-test observation per composition, so what
generalises is the complementary-model-selection principle, not the exact
ensemble.
% Limitations widened (Review 2 "only briefly"; user pick B, middle course):
% seed variance / significance and the domain boundary, not the full pass-1 list.
Every fold model is trained once, so no seed variance is measured and no
significance test backs the differences between compositions.
The internal-to-test transfer is reliable only for full-label
ensembles, so a PCS ensemble's class balance is checked against the prior.
Synthetic augmentation only raises the minority-class counts.
Neither the diversity of the generated tweets nor their value beyond the larger sample size is measured, so the synthetic data may add little and carries generator-specific artefacts.
% Resource limitations sharpened (Philipp, 2026-07-10): name BOTH costs —
% the CV5 pool training (5 fold models per candidate branch, hundreds of
% voters) and the multi-LLM inference (up to three different LLMs per
% ensemble), anchored to the §4.1 figures.
The approach is also resource-hungry.
Every candidate branch fine-tunes five fold models for cross-validation, so hundreds of trained voters stand behind the selection.
At deployment up to three different LLMs cast the nine votes on every tweet (\S\ref{sec:models}).
The strongest internal scores sit on the 128B model, far costlier to fine-tune and run than the compact models, so exploiting its remaining potential hinges on bringing this resource cost down.
Finally, the result is bound to its setting.
The system is built and evaluated on German tweets from one right-wing extremist network, with class definitions that partly follow German criminal law, so neither the compositions nor the class boundaries transfer to another platform, language or legal framework without renewed validation.

\vspace{-3pt}
\section*{Ethics Statement}
The data are German tweets from a network classified as right-wing extremist and
contain harmful content by design. They are used only for detection, moderation
and research under the shared-task licence and the anonymisation tokens are kept with no attempt at de-anonymisation.
Harmful-content classifiers are dual-use.
As with AI systems in other sensitive text domains \citep{steigerwald2026ethics}, their role is assistive, supporting moderation and the analysis of mass communication in investigations, not targeting individuals. All additional data and models are
publicly available under open or Creative-Commons licences.

\vspace{-3pt}
\section*{Acknowledgments}
% Gefördert durch die Deutsche Forschungsgemeinschaft (DFG) – FIP 160 – Projektnummer 549142762
Funded by the Deutsche Forschungsgemeinschaft (DFG, German Research Foundation) -- FIP 160 -- Project-ID 549142762.
All training and inference ran on the GPU cluster of the Center for Artificial Intelligence (KIZ).

\vspace{-3pt}
\section*{Data Availability}
The \nsynth{} synthetic tweets, generation prompt and parameters are released under CC~BY-NC~4.0 at \url{https://github.com/th-nuernberg/nuernberg-nlp-germeval2026}.

% ===========================================================================
\par}

\bibliography{custom}

\appendix
% Appendix-only float relaxation (2026-08-09): the two per-fold tables in one
% column exceed the default \topfraction=0.7, which turned page 13's left
% column into a float-only column and left its lower half EMPTY while the
% (breakable) prompt boxes jumped to the next column. Letting floats take up
% to ~92% of a column puts the box flow back beneath the tables. Scoped here,
% after every main-text float is already placed.
\renewcommand{\topfraction}{0.92}
\renewcommand{\dbltopfraction}{0.9}
\renewcommand{\floatpagefraction}{0.85}
\renewcommand{\textfraction}{0.05}
% ===========================================================================
% (The former "Data and Class Distribution" appendix table is removed: the
% per-subtask labels, counts and augmentation now live in Table~\ref{tab:data}
% in the main text.)
\section{Hyperparameter Details}
\label{sec:app-hparams}
Table~\ref{tab:app-hparams} lists the training hyperparameters shared by all
models between 7 and 14 billion parameters.
The 128B model deviates to keep its cost in bounds, training with LoRA rank 16
($\alpha{=}32$) also for SFT and for three epochs under both methods.
LLaM-I trains for five epochs, as it peaks by epoch three.
The best epoch by validation macro-$\fone$ is deployed in every case.
% Values verified against scripts/train/train_v3_trial.py (SFT) and
% scripts/train/train_clshead.py (ClsHead); the 128B exceptions are config-id 13
% in both scripts. Caption <= 2 lines.
\begin{table}[h]
  \centering
  \small
  \begin{tabular*}{\columnwidth}{@{\extracolsep{\fill}}lcc@{}}
    \toprule
                       & \textbf{SFT} & \textbf{ClsHead} \\
    \midrule
    LoRA rank          & 32   & 16   \\
    LoRA $\alpha$      & 64   & 32   \\
    LoRA dropout       & 0.05 & 0.05 \\
    Learning rate      & $1\mathrm{e}{-4}$ & $2\mathrm{e}{-5}$ \\
    LR schedule        & cosine & cosine \\
    Warmup ratio       & 0.1  & ---  \\
    Weight decay       & 0.01 & 0.01 \\
    Loss               & CE (label) & focal $\gamma{=}2$ \\
    Epochs             & 10   & 10   \\
    Effective batch    & 8    & 8    \\
    \bottomrule
  \end{tabular*}
  \caption{Core training hyperparameters of the 7--14B models (4-bit NF4 QLoRA,
  no truncation). 128B deviations in the text.}
  \label{tab:app-hparams}
\end{table}

\section{Full CV5 Results}
\label{sec:app-matrix}
\label{sec:app-perfold}
% Tables 5-8 replaced the three method-comparison figures (S-panel, G-panel,
% heatmap; make_figures.py still renders them for internal analysis) on
% 2026-07-05: same signal, a fraction of the appendix space. The chosen
% full-label scope (G/S) is named per row.
Tables~\ref{tab:perfold-c2a} to~\ref{tab:perfold-def} list the per-fold $\fcv$ and the top-3-fold mean $\fcvtop$ for every model and method on each subtask, the full-label scope with the best fold coverage in parentheses.
Prompt engineering trails the trained methods on every subtask and, averaged over the models, the gap is widest on DBO and VIO.
\IfFileExists{figures/perfold.tex}{% AUTO-GENERATED by paper/figures/make_tables.py --- do not edit by hand.
% Re-run the script (then ./build.sh -f) to refresh as runs land.

\begin{table}[tp]
  \centering\footnotesize
  \setlength{\tabcolsep}{1.6pt}
  \begin{tabular*}{\columnwidth}{@{\extracolsep{\fill}}llccccc c@{}}
    \toprule
    \textbf{Model} & \textbf{Method} & f0 & f1 & f2 & f3 & f4 & top3 \\
    \midrule
    min & Prompt\,(S) & 62.48 & 61.70 & 62.15 & 62.03 & 62.44 & 62.36 \\
    min & SFT\,(S) & 83.93 & 84.23 & 84.97 & 84.84 & 85.75 & 85.19 \\
    min & ClsHead\,(S) & 86.13 & 85.34 & 86.03 & 84.36 & 85.30 & 85.83 \\
    \addlinespace[1pt]
    phi4 & Prompt\,(S) & 75.59 & 73.61 & 73.67 & 75.24 & 74.75 & 75.19 \\
    phi4 & SFT\,(S) & 85.75 & 86.24 & 86.35 & 85.79 & 85.39 & 86.12 \\
    phi4 & ClsHead\,(G) & 85.85 & 86.49 & 85.84 & 84.48 & 86.24 & 86.19 \\
    \addlinespace[1pt]
    LLaM-B & Prompt\,(S) & 33.13 & 32.96 & 34.38 & 32.81 & 35.17 & 34.23 \\
    LLaM-B & SFT\,(S) & 86.06 & 84.60 & 86.72 & 85.50 & 85.78 & 86.19 \\
    LLaM-B & ClsHead\,(S) & 84.93 & 85.80 & 86.99 & 84.84 & 84.40 & 85.91 \\
    \addlinespace[1pt]
    LLaM-I & Prompt\,(S) & 48.26 & 49.64 & 50.25 & 49.62 & 51.07 & 50.32 \\
    LLaM-I & SFT\,(S) & 86.53 & 84.55 & 86.32 & 85.72 & 84.94 & 86.19 \\
    LLaM-I & ClsHead\,(S) & 85.03 & 87.03 & 85.87 & 85.50 & 84.79 & 86.13 \\
    \addlinespace[1pt]
    Mis128B & Prompt\,(S) & 79.82 & 79.15 & 79.08 & 78.67 & 78.20 & 79.35 \\
    Mis128B & SFT\,(S) & 86.71 & 86.63 & 86.67 & 86.22 & 86.96 & \textbf{86.78} \\
    Mis128B & ClsHead\,(S) & 78.14 & 86.77 & 86.63 & 85.24 & 86.27 & 86.56 \\
    \addlinespace[1pt]
    \bottomrule
  \end{tabular*}
  \caption{C2A: per-fold CV5 macro-$\fone$ (\%), full-label scope in parentheses, best top3 bold.}
  \label{tab:perfold-c2a}
\end{table}

\begin{table}[tp]
  \centering\footnotesize
  \setlength{\tabcolsep}{1.6pt}
  \begin{tabular*}{\columnwidth}{@{\extracolsep{\fill}}llccccc c@{}}
    \toprule
    \textbf{Model} & \textbf{Method} & f0 & f1 & f2 & f3 & f4 & top3 \\
    \midrule
    min & Prompt\,(S) & 18.42 & 18.59 & 18.41 & 17.49 & 17.82 & 18.47 \\
    min & SFT\,(S) & 67.47 & 66.36 & 62.25 & 68.26 & 66.92 & 67.55 \\
    min & ClsHead\,(S) & 66.11 & 68.80 & 65.57 & 66.49 & 64.95 & 67.13 \\
    \addlinespace[1pt]
    phi4 & Prompt\,(S) & 33.91 & 33.63 & 36.39 & 33.17 & 31.85 & 34.64 \\
    phi4 & SFT\,(S) & 64.15 & 70.87 & 65.20 & 63.46 & 55.30 & 66.74 \\
    phi4 & ClsHead\,(S) & 69.20 & 68.50 & 66.55 & 61.77 & 67.54 & 68.41 \\
    \addlinespace[1pt]
    LLaM-B & Prompt\,(S) & 22.41 & 23.06 & 21.93 & 22.64 & 22.15 & 22.70 \\
    LLaM-B & SFT\,(S) & 62.82 & 65.87 & 57.28 & 65.05 & 59.78 & 64.58 \\
    LLaM-B & ClsHead\,(S) & 59.66 & 69.33 & 68.43 & 58.97 & 64.85 & 67.54 \\
    \addlinespace[1pt]
    LLaM-I & Prompt\,(S) & 22.42 & 22.19 & 22.49 & 24.01 & 22.27 & 22.98 \\
    LLaM-I & SFT\,(S) & 67.70 & 64.53 & 65.25 & 70.66 & 55.94 & 67.87 \\
    LLaM-I & ClsHead\,(S) & 60.24 & 66.69 & 66.01 & 64.70 & 66.85 & 66.52 \\
    \addlinespace[1pt]
    Mis128B & Prompt\,(S) & 35.39 & 34.11 & 35.93 & 33.20 & 32.47 & 35.14 \\
    Mis128B & SFT\,(G) & 69.49 & 71.00 & 68.21 & 71.83 & 60.29 & \textbf{70.77} \\
    Mis128B & ClsHead\,(S) & 59.02 & 71.15 & 24.08 & 66.07 & 60.57 & 65.93 \\
    \addlinespace[1pt]
    \bottomrule
  \end{tabular*}
  \caption{DBO: per-fold CV5 macro-$\fone$ (\%), full-label scope in parentheses, best top3 bold.}
  \label{tab:perfold-dbo}
\end{table}

\begin{table}[tp]
  \centering\footnotesize
  \setlength{\tabcolsep}{1.6pt}
  \begin{tabular*}{\columnwidth}{@{\extracolsep{\fill}}llccccc c@{}}
    \toprule
    \textbf{Model} & \textbf{Method} & f0 & f1 & f2 & f3 & f4 & top3 \\
    \midrule
    min & Prompt\,(S) & 33.56 & 35.73 & 27.08 & 31.50 & 34.29 & 34.53 \\
    min & SFT\,(S) & 45.75 & 54.93 & 45.26 & 50.31 & 53.01 & 52.75 \\
    min & ClsHead\,(G) & 48.98 & 55.96 & 54.46 & 47.66 & 56.87 & 55.76 \\
    \addlinespace[1pt]
    phi4 & Prompt\,(S) & 19.72 & 20.32 & 17.90 & 20.92 & 21.95 & 21.06 \\
    phi4 & SFT\,(S) & 48.73 & 55.16 & 43.78 & 41.79 & 61.07 & 54.98 \\
    phi4 & ClsHead\,(S) & 51.96 & 53.75 & 49.14 & 48.27 & 55.60 & 53.77 \\
    \addlinespace[1pt]
    LLaM-B & Prompt\,(S) & 10.54 & 10.66 & 10.61 & 10.69 & 10.42 & 10.65 \\
    LLaM-B & SFT\,(S) & 54.62 & 51.35 & 45.71 & 44.40 & 54.71 & 53.56 \\
    LLaM-B & ClsHead\,(G) & 52.39 & 54.28 & 48.61 & 48.59 & 51.67 & 52.78 \\
    \addlinespace[1pt]
    LLaM-I & Prompt\,(S) & 15.87 & 15.46 & 16.23 & 15.46 & 14.94 & 15.85 \\
    LLaM-I & SFT\,(S) & 50.28 & 51.25 & 45.77 & 45.69 & 53.58 & 51.70 \\
    LLaM-I & ClsHead\,(G) & 50.01 & 53.07 & 40.37 & 44.43 & 49.00 & 50.70 \\
    \addlinespace[1pt]
    Mis128B & Prompt\,(S) & 34.04 & 35.47 & 34.36 & 38.23 & 32.03 & 36.02 \\
    Mis128B & SFT\,(S) & 47.04 & 62.56 & 53.96 & 47.01 & 61.09 & \textbf{59.21} \\
    Mis128B & ClsHead\,(S) & 49.32 & 58.48 & 58.52 & 46.44 & 51.49 & 56.17 \\
    \addlinespace[1pt]
    \bottomrule
  \end{tabular*}
  \caption{VIO: per-fold CV5 macro-$\fone$ (\%), full-label scope in parentheses, best top3 bold.}
  \label{tab:perfold-vio}
\end{table}

\begin{table}[tp]
  \centering\footnotesize
  \setlength{\tabcolsep}{1.6pt}
  \begin{tabular*}{\columnwidth}{@{\extracolsep{\fill}}llccccc c@{}}
    \toprule
    \textbf{Model} & \textbf{Method} & f0 & f1 & f2 & f3 & f4 & top3 \\
    \midrule
    min & Prompt\,(S) & 50.83 & 52.95 & 51.98 & 51.26 & 51.98 & 52.30 \\
    min & SFT\,(S) & 76.09 & 80.61 & 79.12 & 77.57 & 82.34 & 80.69 \\
    min & ClsHead\,(S) & 79.18 & 79.12 & 80.90 & 82.50 & 82.44 & 81.95 \\
    \addlinespace[1pt]
    phi4 & Prompt\,(S) & 67.01 & 67.14 & 69.23 & 66.01 & 66.59 & 67.80 \\
    phi4 & SFT\,(S) & 79.41 & 78.88 & 80.56 & 81.03 & 82.52 & 81.37 \\
    phi4 & ClsHead\,(S) & 78.37 & 78.67 & 80.02 & 83.09 & 82.90 & 82.01 \\
    \addlinespace[1pt]
    LLaM-B & Prompt\,(S) & 44.94 & 45.49 & 44.09 & 50.30 & 42.36 & 46.91 \\
    LLaM-B & SFT\,(S) & 79.17 & 79.84 & 79.34 & 82.93 & 82.98 & 81.92 \\
    LLaM-B & ClsHead\,(S) & 79.35 & 80.06 & 79.04 & 83.87 & 81.33 & 81.75 \\
    \addlinespace[1pt]
    LLaM-I & Prompt\,(S) & 40.72 & 41.44 & 40.97 & 38.16 & 42.77 & 41.73 \\
    LLaM-I & SFT\,(S) & 78.65 & 76.00 & 78.51 & 82.91 & 82.41 & 81.33 \\
    LLaM-I & ClsHead\,(S) & 79.47 & 79.57 & 77.92 & 82.20 & 82.40 & 81.39 \\
    \addlinespace[1pt]
    Mis128B & Prompt\,(S) & 68.86 & 66.43 & 73.55 & 70.85 & 67.09 & 71.09 \\
    Mis128B & SFT\,(S) & 77.23 & 83.70 & 78.87 & 84.27 & 83.40 & \textbf{83.79} \\
    Mis128B & ClsHead\,(S) & 79.73 & 76.32 & 78.40 & 83.99 & 79.70 & 81.14 \\
    \addlinespace[1pt]
    \bottomrule
  \end{tabular*}
  \caption{DEF: per-fold CV5 macro-$\fone$ (\%), full-label scope in parentheses, best top3 bold.}
  \label{tab:perfold-def}
\end{table}
}{%
  \par\TODO{run \texttt{python paper/figures/make\_tables.py} to generate
  \texttt{figures/perfold.tex}.}}

% ===========================================================================
\section{Embedding Geometry}
\label{sec:app-emb}
\begin{figure*}[t]
  \centering
  \includegraphics[width=\textwidth]{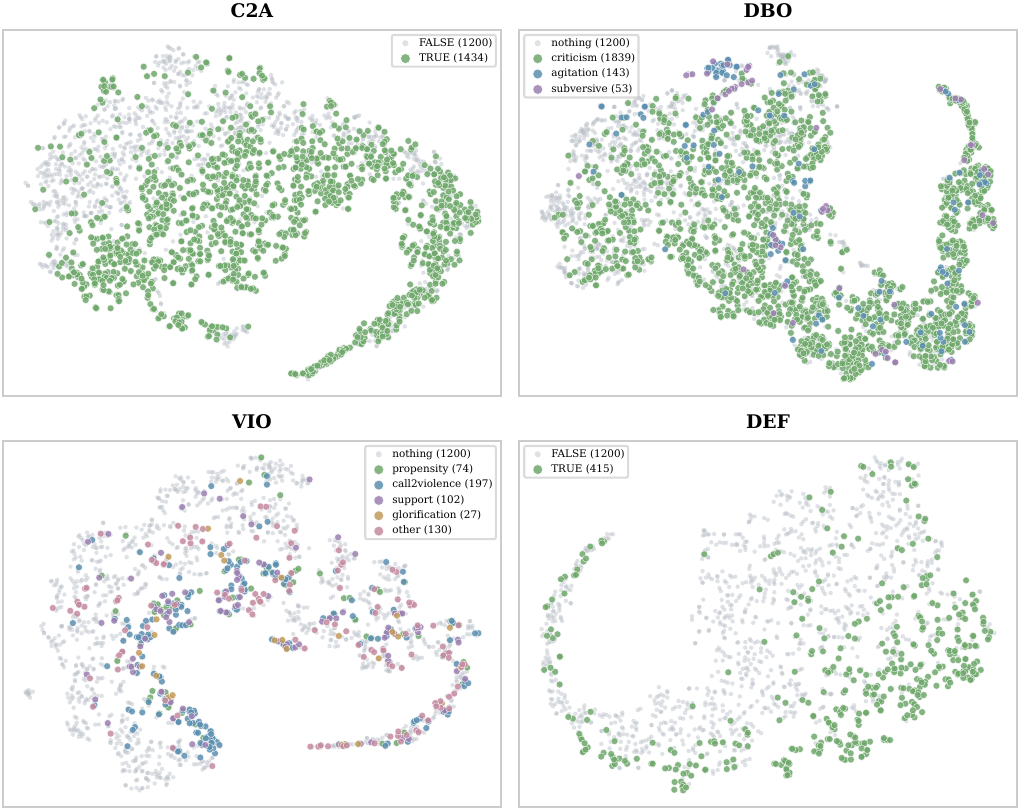}
  \caption{Per-subtask t-SNE of the untrained 128B last-token hidden states,
  with the rare positive classes overlapping off the shelf.}
  \label{fig:emb}
\end{figure*}

% Positive-class cap removed 2026-07-10 (Philipp): every positive instance is
% plotted, only the majority class is sampled (1,200 per subtask).
Figure~\ref{fig:emb} shows a t-SNE of the untrained 128B model's last-token hidden
states per subtask, every positive instance kept and the majority class sampled to 1{,}200 (legends).
In this raw representation the rare positive classes sit close together with no
clean boundary between them, so the strongest model in the pool cannot separate
the decisive classes off the shelf.
This is the geometric motivation for the positive-class specialist and it mirrors why prompt engineering plateaus where fine-tuning breaks through.
% Figure regenerated 2026-07-01 from the landed mis128-base-fulltrain npz files
% (job 179424) via paper/figures/make_embeddings_fig.py; reading confirmed.

% ===========================================================================
\section{Prompts}
\label{sec:app-prompts}
% SFT prompts verbatim from scripts/config.py (SUBTASK_SYSTEM_PROMPTS +
% SUBTASK_CATEGORY_DESCRIPTIONS). Each box: German original first, English
% translation in italics below the dashed rule (\tcblower). The boxes are
% INLINE and breakable, so they flow with the text across columns/pages and
% leave no float whitespace. (The old "mis-break" warning applied to the
% sidebyside variant, which cannot break; stacked DE/EN breaks fine.)
Every SFT voter uses the same template --- the subtask's system prompt, the
tweet, the class definitions under a \textit{\#\# Kategorien} header and the
label as the assistant answer.
One box per subtask below lists this system prompt and the class definitions,
the German original first and the English translation in italics below the
dashed rule.

% Shared style for the four prompt boxes.
\tcbset{promptbox/.style={
  breakable, colback=clrPrompt!8!white, colframe=clrPrompt!70!black,
  fontupper=\fontsize{7.5}{9}\selectfont, fontlower=\fontsize{7.5}{9}\selectfont,
  boxrule=0.5pt, left=4pt, right=4pt, top=3pt, bottom=3pt,
  toptitle=1.5pt, bottomtitle=1.5pt,
  segmentation style={draw=black!45, dashed, line width=0.5pt},
}}

\begin{tcolorbox}[promptbox,
  title={\fontsize{8}{9.5}\selectfont\bfseries C2A SFT prompt},
]
\textbf{System:} Du bist ein Experte f\"ur die Erkennung von Aufrufen
zu Handlungen in deutschsprachigen Social-Media-Beitr\"agen. Antworte
ausschlie\ss{}lich mit dem Label.\\[1.5pt]
\textbf{Kategorien:}\\
\texttt{FALSE}: Der Tweet enth\"alt keinen Aufruf zu einer bestimmten Handlung
oder einem bestimmten Verhalten. Allgemeine Meinungs\"au\ss{}erungen,
Beschreibungen oder Kommentare ohne Handlungsaufforderung fallen in diese
Kategorie.\\
\texttt{TRUE}: Der Tweet enth\"alt einen Aufruf zu einer bestimmten Handlung
oder einem bestimmten Verhalten (Call to Action). Die Handlung muss nicht
strafrechtlich relevant sein --- auch Aufrufe zu Demonstrationen, politischen
Kampagnen, Flugblattverteilungen, Boykotten oder \"ahnlichen Aktionen z\"ahlen.
Entscheidend ist, dass eine konkrete Aufforderung an den Leser gerichtet wird,
etwas Bestimmtes zu tun.
\tcblower
{\itshape \textbf{System:} You are an expert in detecting calls to action in
German-language social-media posts. Respond with the label only.\\[1.5pt]
\textbf{Categories:}\\
\texttt{FALSE}: the tweet contains no call to a specific action or behaviour;
general opinions, descriptions or comments without a request to act.\\
\texttt{TRUE}: the tweet contains a call to a specific action or behaviour;
the action need not be criminal --- calls to demonstrations, political
campaigns, leafleting or boycotts count, decisive is a concrete request to the
reader to do something specific.}
\end{tcolorbox}

\begin{tcolorbox}[promptbox,
  title={\fontsize{8}{9.5}\selectfont\bfseries DBO SFT prompt},
]
\textbf{System:} Du bist ein Experte f\"ur die Erkennung von Angriffen
auf die freiheitlich-demokratische Grundordnung der Bundesrepublik Deutschland.
Antworte ausschlie\ss{}lich mit dem Label.\\[1.5pt]
\textbf{Kategorien:}\\
\texttt{nothing}: Der Tweet enth\"alt weder Kritik noch einen Angriff auf die
freiheitliche demokratische Grundordnung der Bundesrepublik Deutschland.
Neutrale oder positive \"Au\ss{}erungen zu Regierungsentscheidungen fallen
ebenfalls in diese Kategorie.\\
\texttt{criticism}: Der Tweet enth\"alt berechtigte, legale Kritik an der
Regierung, an Beamten, Beh\"orden oder politischen Parteien. Kritik ist
sachlich und greift keine demokratischen Grundstrukturen an.\\
\texttt{agitation}: Der Tweet enth\"alt agitatorische Bestrebungen. Dazu
geh\"oren die Ank\"undigung von Aktionen wie das Verbreiten von
Propagandamaterial verfassungswidriger oder terroristischer Organisationen oder
die Sch\"andung staatlicher Symbole wie der Flagge der Bundesrepublik
(\S86 StGB, \S90a StGB).\\
\texttt{subversive}: Im Tweet wird der Wille ge\"au\ss{}ert, die bestehende
Regierung gewaltsam zu beseitigen und zu st\"urzen, zum Beispiel durch
militante Aktionen, St\"orung der Stromversorgung oder andere subversive
Ma\ss{}nahmen.
\tcblower
{\itshape \textbf{System:} You are an expert in detecting attacks on the free
democratic basic order of the Federal Republic of Germany. Respond with the
label only.\\[1.5pt]
\textbf{Categories:}\\
\texttt{nothing}: neither criticism nor an attack on the free democratic basic
order, incl.\ neutral or positive statements on government decisions.\\
\texttt{criticism}: legitimate, legal, factual criticism of the government,
officials, authorities or parties that does not attack democratic structures.\\
\texttt{agitation}: agitatory efforts such as announcing the spread of
propaganda of unconstitutional or terrorist organisations or desecrating state
symbols like the federal flag (\S86, \S90a StGB).\\
\texttt{subversive}: the will to violently remove and overthrow the
government, e.g.\ through militant actions or disrupting the power supply.}
\end{tcolorbox}

\begin{tcolorbox}[promptbox,
  title={\fontsize{8}{9.5}\selectfont\bfseries VIO SFT prompt},
]
\textbf{System:} Du bist ein Experte f\"ur die Erkennung von
gewaltbezogenen Aussagen in deutschsprachigen Social-Media-Beitr\"agen.
Antworte ausschlie\ss{}lich mit dem Label.\\[1.5pt]
\textbf{Kategorien:}\\
\texttt{nothing}: Keine gewaltbezogene \"Au\ss{}erung. Der Tweet enth\"alt
keinerlei negative oder besorgniserregende Aussagen zu Gewalt.\\
\texttt{prospensity}: Gewaltbereitschaft. Der Verfasser \"au\ss{}ert den Willen
oder Wunsch, selbst Gewalt anzuwenden. Es geht um die eigene Bereitschaft zur
Gewaltaus\"ubung, nicht um einen Aufruf an andere.\\
\texttt{call2violence}: Aufruf zur Gewalt. Der Verfasser ruft andere Personen
dazu auf oder stachelt sie an, eine Gewalttat zu begehen. Der Unterschied zu
prospensity: hier wird die Gewalt nicht selbst angek\"undigt, sondern andere
werden aufgefordert.\\
\texttt{support}: Billigung von Gewalt. Der Verfasser \"au\ss{}ert positive
Zustimmung zu Gewalt oder einem gewaltt\"atigen Ereignis. Er bef\"urwortet die
Tat, ohne sie selbst zu verherrlichen.\\
\texttt{glorification}: Verherrlichung von Gewalt. Gewalt wird als etwas
besonders Ruhmreiches und Heldenhaftes dargestellt --- st\"arker als blo\ss{}e
Billigung (support). Die Gewalttat oder ihre Ausf\"uhrenden werden
glorifiziert.\\
\texttt{other}: Sonstige besorgniserregende gewaltbezogene \"Au\ss{}erungen,
die in keine der obigen Kategorien passen, aber dennoch einen beunruhigenden
Gewaltbezug aufweisen.
\tcblower
{\itshape \textbf{System:} You are an expert in detecting violence-related
statements in German-language social-media posts. Respond with the label
only.\\[1.5pt]
\textbf{Categories:}\\
\texttt{nothing}: no violence-related statement whatsoever.\\
\texttt{prospensity}: readiness for violence --- the author expresses the will
or wish to use violence themselves, not a call on others.\\
\texttt{call2violence}: the author calls on or incites others to commit a
violent act.\\
\texttt{support}: approval of violence or a violent event without glorifying
it.\\
\texttt{glorification}: violence portrayed as particularly glorious and
heroic, stronger than mere approval.\\
\texttt{other}: other worrying violence-related statements that fit none of
the above categories.}
\end{tcolorbox}

\begin{tcolorbox}[promptbox,
  title={\fontsize{8}{9.5}\selectfont\bfseries DEF SFT prompt},
]
\textbf{System:} Du bist ein Experte f\"ur die Erkennung von
Beleidigung und Verleumdung nach \S\S185-187 StGB in deutschsprachigen
Social-Media-Beitr\"agen. Antworte ausschlie\ss{}lich mit dem Label.\\[1.5pt]
\textbf{Kategorien:}\\
\texttt{FALSE}: Der Tweet enth\"alt keine Beleidigung, \"uble Nachrede oder
Verleumdung im Sinne der \S\S185-187 des deutschen Strafgesetzbuches (StGB).
Sachliche Kritik, allgemeine negative Meinungen oder unh\"ofliche Sprache ohne
konkreten Personenbezug fallen nicht unter Beleidigung im strafrechtlichen
Sinne.\\
\texttt{TRUE}: Der Tweet erf\"ullt den Tatbestand einer ehrverletzenden
Straftat nach \S\S185-187 StGB. Dazu geh\"oren: Beleidigung (\S185) ---
herabsetzende \"Au\ss{}erungen gegen eine bestimmte Person; \"Uble Nachrede
(\S186) --- Behauptung ehrenr\"uhriger Tatsachen, die nicht erweislich wahr
sind; Verleumdung (\S187) --- wissentlich unwahre ehrenr\"uhrige
Tatsachenbehauptungen. Entscheidend ist der strafrechtliche Ma\ss{}stab, nicht
das subjektive Empfinden.
\tcblower
{\itshape \textbf{System:} You are an expert in detecting insult and
defamation under \S\S185--187 of the German Criminal Code in German-language
social-media posts. Respond with the label only.\\[1.5pt]
\textbf{Categories:}\\
\texttt{FALSE}: no insult, malicious gossip or defamation under \S\S185--187
StGB; factual criticism, general negative opinions or impolite language
without reference to a specific person do not count.\\
\texttt{TRUE}: the tweet meets the elements of an honour offence under
\S\S185--187 StGB: insult (\S185) --- derogatory statements against a specific
person; malicious gossip (\S186) --- dishonouring factual claims not provably
true; defamation (\S187) --- knowingly false dishonouring claims. The
criminal-law standard is decisive, not subjective feeling.}
\end{tcolorbox}

% ===========================================================================
\section{Augmentation Prompt}
\label{sec:app-augprompt}
Every synthetic tweet was produced with the batched GPT-5.4 prompt below,
filled per target class with the subtask name and description, the class label
and its definition, the batch size and five randomly sampled real tweets of
that class. Placeholders are in \texttt{\{braces\}}; the German original is
shown with an English translation.

\begin{tcolorbox}[promptbox,
  title={\fontsize{8}{9.5}\selectfont\bfseries Augmentation prompt (GPT-5.4, batch API)},
]
\textbf{System:} Du bist ein Experte f\"ur die Analyse rechtsextremer Inhalte
in deutschen Social-Media-Beitr\"agen. Generiere realistische
Trainingsbeispiele f\"ur ein Klassifikationssystem. Gib nur valides JSON
aus.\\[2pt]
\textbf{User:}\\
\#\# Hintergrund\\
Die Daten stammen aus einem rechtsextremen Twitter-Netzwerk (\"offentliche
deutschsprachige Posts und Kommentare, Dezember 2014 bis Juli 2016).
Erw\"ahnungen sind anonymisiert: [@IND] = Einzelperson, [@GRP] =
Gruppe/Organisation, [@POL] = Polizei/Beh\"orde, [@PRE] = Presse/Medien.\\[1.5pt]
\#\# Aufgabe: \texttt{\{subtask\}}\quad \texttt{\{subtask\_description\}}\\
\#\# Ziel-Kategorie: \texttt{\{label\}}\quad \texttt{\{label\_definition\}}\\[1.5pt]
\#\# Anforderungen\\
1. Generiere \texttt{\{n\}} NEUE, DIVERSE deutsche Tweets der Kategorie
``\texttt{\{label\}}''.\\
2. Die Tweets m\"ussen REALISTISCH klingen (rechtsextremes Netzwerk,
2014--2016).\\
3. VARIIERE die Themen: Regierungskritik, Migration, Islam, politische Gegner,
gesellschaftliche Konflikte.\\
4. Verwende nat\"urliches, umgangssprachliches Deutsch.\\
5. Nutze gelegentlich die Anonymisierungstokens [@IND], [@GRP], [@POL],
[@PRE].\\
6. Jeder Tweet maximal 280 Zeichen.\\
7. Die Tweets m\"ussen EINDEUTIG die Kategorie ``\texttt{\{label\}}''
widerspiegeln.\\[1.5pt]
\#\# Beispiele aus echten Trainingsdaten\\
\texttt{\{5 real examples of the target class\}}\\[1.5pt]
\#\# Output\\
Generiere genau \texttt{\{n\}} Tweets als JSON-Array; nur valides JSON, keine
Erkl\"arungen.\\[3pt]
{\itshape\textbf{System (EN):} You are an expert in analysing right-wing
extremist content in German social-media posts. Generate realistic training
examples for a classification system. Output valid JSON only.\\[2pt]
\textbf{User (EN):} \#\# Background --- the data are German tweets from a
right-wing extremist network (public posts, Dec.\ 2014--Jul.\ 2016), with
mentions anonymised ([@IND] individual, [@GRP] group, [@POL] police, [@PRE]
press). \#\# Task / target category are filled per class with the subtask and
label definitions. \#\# Requirements: generate \texttt{\{n\}} new, diverse,
realistic German tweets of the target class; vary the topics; natural
colloquial German; occasional anonymisation tokens; at most 280 characters;
each tweet must clearly reflect the target class. \#\# Examples: five real
tweets of that class. \#\# Output: exactly \texttt{\{n\}} tweets as a JSON
array, valid JSON only.}
\end{tcolorbox}

\end{document}